\documentclass[letterpaper, 10 pt, conference]{ieeeconf}

\IEEEoverridecommandlockouts % This command is only needed if you want to use the \thanks command

\usepackage{amsmath,amsfonts}
\usepackage{algorithmic}
\usepackage{algorithm}
\usepackage{array}
\usepackage[caption=false,font=normalsize,labelfont=sf,textfont=sf]{subfig}
\usepackage{textcomp}
\usepackage{stfloats}
\usepackage{url}
\usepackage{verbatim}
\usepackage{graphicx}
\usepackage{cite}
\usepackage{booktabs}
\usepackage{multirow}
\title{\LARGE \bf
ERASOR++: Height Coding Plus Egocentric Ratio Based Dynamic Object Removal for Static Point Cloud Mapping}

\author{Jiabao Zhang\textsuperscript{1} and Yu Zhang\textsuperscript{1}\textsuperscript{2}$^{\ast}$
\thanks{\textsuperscript{1}State Key Laboratory of Industrial Control Technology, College of Control Science and Engineering, Zhejiang University, Hangzhou, China.}
\thanks{\textsuperscript{2}Key Laboratory of Collaborative Sensing and Autonomous Unmanned Systems of Zhejiang Province, Hangzhou, China.}
\thanks{$^{\ast}$Corresponding author: Yu Zhang, {\tt\small zhangyu80@zju.edu.cn}}
}

\begin{document}

\maketitle
\pagestyle{empty}  % no page number for the second and the later pages
\thispagestyle{empty} % no page number for the first page

\begin{abstract}
Mapping plays a crucial role in location and navigation within automatic systems. However, the presence of dynamic objects in 3D point cloud maps generated from scan sensors can introduce map distortion and long traces, thereby posing challenges for accurate mapping and navigation. 
To address this issue, we propose ERASOR++, an enhanced approach based on the Egocentric Ratio of Pseudo Occupancy for effective dynamic object removal.
% that incorporates the height coding descriptor, addition test methods, and the Egocentric Ratio of Pseudo Occupancy based methods for effective dynamic object removal.
To begin, we introduce the Height Coding Descriptor, which combines height difference and height layer information to encode the point cloud. Subsequently, we propose the Height Stack Test, Ground Layer Test, and Surrounding Point Test methods to precisely and efficiently identify the dynamic bins within point cloud bins, thus overcoming the limitations of prior approaches.
Through extensive evaluation on open-source datasets, our approach demonstrates superior performance in terms of precision and efficiency compared to existing methods. Furthermore, the techniques described in our work hold promise for addressing various challenging tasks or aspects through subsequent migration.

%Mapping plays an important role in location and navigation in automatic systems. As for 3D point cloud maps generated from scan sensors, there may be map distortion and long traces caused by dynamic objects, which makes dynamic removal a worthy problem.
%Therefore, in our work, we proposed ERASOR++, height coding descriptor and addition methods plus Egocentric Ratio of Pseudo Occupancy based dynamic object removal. 
%First, the  Height Coding Descriptor which combines the height difference and height layer information is designed to represent point cloud occupancy from scan and submap. Then, the Height Stack Test, along with Ground Point and Surrounding Point Test methods are proposed to accurately and effectively fetch the dynamic bin from all point cloud bins, which successfully solves the limitations of previous work.
%Finally, after testing the full system on open-source datasets, it proved that our proposed work outperforms the previous work on both precision and efficiency. And the methods in our work may contribute to subsequent migration in other challenging tasks or aspects.
\end{abstract}

% \begin{IEEEkeywords}
% Dynamic Object Removal, Mapping, Height Coding, LiDAR Descriptor.
% \end{IEEEkeywords}

\section{Introduction}
Map building is an essential part of numerous automatic systems, including autonomous cars and unmanned ground or aerial vehicles, among others, enabling accurate location and navigation. Various map representations are available based on the location system, encompassing feature-based maps, metric maps, and semantic maps \cite{global} \cite{semantic}.
This paper specifically concentrates on metric maps within the context of 3D point cloud maps, which are typically derived from laser scans, especially LiDAR sensors.
% such as autonomous cars, unmanned ground or aerial vehicles, and so on, utilized for location and navigation. According to the location system, there are various kinds of map representation including feature-based maps, metric maps and semantic maps \cite{global} \cite{semantic}.
% This paper is focused on one category of metric maps, 3D point cloud map, which is always obtained from laser scans, especially LiDAR sensors.(and prominently employ LiDAR sensors.)

For accurate location and navigation, the creation of maps within static environments is considered ideal, but the existence of dynamic objects is inevitable in the majority of real-world scenarios.
Since each scan point cloud captures data at discrete time intervals, the presence of dynamic objects can introduce distortions and prolonged traces within the point cloud data, thereby negatively affecting the mapping process \cite{erasor}.
% As there is a time interval between each point or frame in each scan point cloud, dynamic objects may cause distortion and long traces in point cloud data that have an adverse impact on mapping \cite{erasor}.

% To deal with the problem mentioned above, many dynamic object removal methods for LiDAR point cloud are proposed, 
To address the aforementioned problem, some methods have been proposed to remove dynamic objects from LiDAR point clouds, which can be generally divided into online methods applied during the map generation process \cite{online} \cite{dynamicfilter}, and the post-processing methods implemented after gaining the generated map. 
% 0816 add the motivation for studying the offline method
% On account of achieving the instantaneuty with only a few current scans, the online methods are always limited to detecting objects moving rapidly in the current scene \cite{online}, which may lose sight of some traces of moving objects. And the accuracy of online methods are generally lower than post-processing methods even with scan-map front end and map-map back end \cite{dynamicfilter}.
Due to their ability to achieve near-instantaneity using a limited number of current scans, online methods have inherent limitations in detecting rapidly moving objects within the present scene \cite{online}, which may overlook certain traces of moving objects. And despite employing scan-map front end and map-map back end techniques, online methods generally exhibit lower accuracy compared to post-processing methods \cite{dynamicfilter}.

For better accuracy, we concern with the latter category, post-processing methods, which includes cluster-based methods such as \cite{training} \cite{pointlearn}, semantic-based methods such as \cite{rangenet} \cite{recurrentoct}, voxel-based (or ray-tracing based) methods such as \cite{2020robust} \cite{peopleremover} \cite{octomap}, visibility-based methods such as \cite{remove} \cite{longterm}, and descriptor-based methods \cite{erasor}.
Each of these methods offers a unique perspective on addressing the challenge of dynamic object removal from LiDAR point clouds.

% \IEEEpubidadjcol

Despite their contributions, these methods have certain limitations that can be found in Section \ref{RelatedWorks} in detail. 
These limitations underscore the need for further advancements in dynamic object removal techniques for LiDAR point clouds.

\begin{figure*}[!t]
\centering
\includegraphics[width=7in]{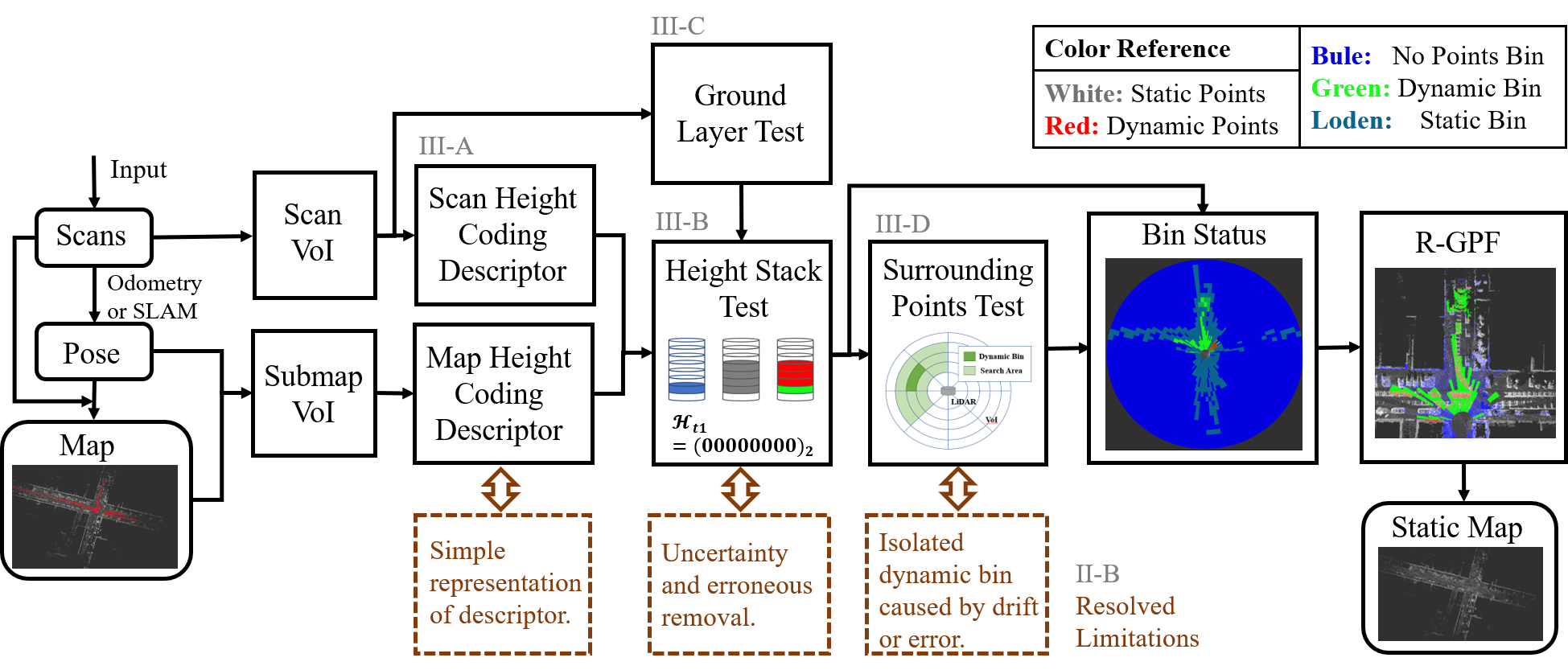}
\caption{Framework of ERASOR++. The main parts are labeled with corresponding sections, with respective resolved limitations mentioned in related works. 
Note that the processes before the VoI part and after the R-GPF part transfer from the previous work ERASOR\cite{erasor}.}
\label{0Framework}
\end{figure*}

In this paper of work, we follow the thought of descriptor-based method ERASOR \cite{erasor}, introduce a novel representation of descriptors, and multiple effective test methods to overcome existing limitations. The framework of our proposed methods is depicted in Fig.\ref{0Framework}.
% The framework of our proposed additional part along with the reference part is depicted in Fig.\ref{0Framework}.

The main contributions of our paper are summarized as follows:

\begin{itemize}
\item{{\bf{Novel combination and representation of Height Coding Descriptor (HCD),}} 
% We proposed the Height Coding Descriptor (HCD), 
which contains both extreme values and sparse middle values of the Z axis, which enhances the descriptive capabilities.
%We combined the height difference and height layer encoding information and proposed the Height Coding Descriptor (HCD), containing both extreme value and sparse middle values of the Z axis, which enhances the descriptive capabilities.
}
\item{{\bf{Comprehensive Height Stack Test (HST) method for evaluating dynamic bins,}} 
% We proposed the Height Stack Test (HST) based on the HCD, 
which leverages the description of points in the middle height area in HCD and could avoid potential bad dynamic removal issues associated with inadequate information.
% using the description of points in the middle height area, which could avoid potential bad dynamic removal issues associated with inadequate information regarding point cloud overlap areas along the height direction.
% which could avoid bad dynamic removal caused by a lack of information about point cloud overlap areas in the height direction.
}
\item{{\bf{Additional Ground Layer Test (GLT) and Surrounding Points Test (SPT) for accuracy of dynamic status,}} 
% Two additional tests were proposed, 
% for the first could check the ground layer in HCD to provide the relative position for HST, and the second cloud correct the isolated bin status by searching points around, 
which both serve to complement the algorithm and address specific challenges.
}
\item{{\bf{Superior performance compared with the previous work.}} We compared our proposed ERASOR++ method with the previous ERASOR \cite{erasor} method, showing that our proposed method outperforms the previous work on both precision and efficiency.}
\end{itemize}

The rest of this paper contains Related Works indicated in Section \ref{RelatedWorks}, Methodology in ERASOR++ described in Section \ref{Methodology}, Experiments discussed in Section \ref{Experiments}, as well as Conclusion summarized in Section \ref{Conclusion}.

\section{Related Works}\label{RelatedWorks}
% In this section, we summarize related works about dynamic object removal for static 3D point cloud maps, focusing on the limitations of descriptor based method, and generalize literature related to LiDAR descriptor.

%focused on the main content in Egocentric RAtio of pSeudo Occupancy-based dynamic object Removal (ERASOR) \cite{erasor} method, proposed by Lim \emph{et al.}. 

\subsection{Dynamic Object Removal Methods}
% \textbf{Dynamic Object Removal Methods.} 
Categories of typical post-processing methods are summarized as follows. 
Cluster-based \cite{training} \cite{pointlearn} or semantic-based \cite{rangenet} \cite{recurrentoct} methods, which often rely on deep learning networks \cite{learning}. These methods typically require point cloud supervised labels for training, which can limit their generalization ability. 
Voxel-based methods \cite{2020robust} \cite{peopleremover} \cite{octomap} tend to be computationally expensive due to the calculation of misses or hits within each voxel. 
Visibility-based methods \cite{remove} \cite{longterm} heavily depend on the hypothesis that dynamic objects always appear in front of static objects along a ray, rendering them invalid in scenarios where obstacles obstruct the line of sight between sensors and dynamic objects, or in open environments with few regular static objects.
% However, some limitations can be found in these methods. 
% The cluster or semantic based methods mostly using deep learning networks \cite{learning} may rely on the point cloud supervised label and lack of generalization ability. 
% The voxel based methods are always computationally expensive caused by calculating the misses or hits times in each voxel. 
% The visibility based methods strongly depend on the hypothesis that dynamic objects always appear in front of static objects along one ray, which makes it invalid when obstacles appear between sensors and dynamic objects or when in an open environment without many regular static objects. 
 
As for the descriptor-based method like Egocentric RAtio of pSeudo Occupancy-based dynamic object Removal (ERASOR) \cite{erasor} method, proposed by Lim \emph{et al.}, it leverages vertical column descriptors and could overcome the limitations of above with high running speed and a visibility-free characteristic.
% , which seems a great solution for dynamic object removal in static map building. 
But it still has some problems like bad removal in some areas, especially when used in unstructured or vegetational environments as shown in the next part. 
 % Furthermore, the main work and limitations of this method in detail can be found in Section \ref{RelatedWorks}. 

\subsection{Main Works and Limitations in ERASOR}\label{limitationslist}
ERASOR method focuses on the problem of dynamic objects post-rejection in a set of generated maps.
% , which can be classified into segmentation-based, ray tracing-based, visibility-based and egocentric descriptor-based.
In that paper, Lim \emph{et al.} brought egocentric descriptor into dynamic area test and proposed the major work using egocentric descriptor-based method. 
% The main framework of ERASOR is summarized as follows.
The main work of ERASOR includes leveraging a representation of points in vertical columns called Region-wise Pseudo Occupancy Descriptor (R-POD), a method called Scan Ratio Test (SRT) to fetch bins with dynamic objects, a static point retrieval method called Region-wise Ground Plane Fitting (R-GPF), and presenting a metrics, especially for static map building tasks.
% The major framework of ERASOR is shown in Fig.\ref{0Framework}. 
% Getting scan data, pose estimation and prior map as input, ERASOR method first encodes query scan point cloud and its corresponding submap point cloud into Volume of Interest (VoI), through which the points in query area are fetched and divided into different bins according to the horizontal position. Then the representation of points in vertical columns called Region-wise Pseudo Occupancy Descriptor (R-POD) and the testing method called Scan Ratio Test (SRT) are leveraged to get bin status, describing whether there are dynamic objects in each bin. Finally, with an attention to the characteristics of most dynamic objects in urban settings being in contact with the ground, the static point retrieval method called Region-wise Ground Plane Fitting (R-GPF) is employed within each dynamic bin to fit ground points and eliminate the dynamic objects above ground.
More details about the mentioned work can be found in \cite{erasor}.

%\subsection{Limitations in ERASOR}
 Although ERASOR provides a promising performance on dynamic object removal, it still has some limitations, which can be summarized as follows. 
 % and shown in Fig.\ref{0limitation}.

\begin{itemize}
\item{{\bf{Simple representation of descriptor.}} In the previous method, the representation of the bin area's descriptor was simplified to only include the height difference along the Z-axis, which seems a substantial loss of valuable information.
% between the maximum and minimum values along the Z-axis.
% the used information of bin area in representation is just the height difference between the maximum value and the minimum value in the Z axis. 
% Given the representation as main elements, relying solely on extreme values seems a substantial loss of valuable information.
% As main elements of scan descriptors, directly utilizing extreme value seems a great loss of other points' information.
}
\item{{\bf{Uncertainty and erroneous removal in areas with blocked ground.}} As shown in Fig.\ref{2StackRatio}, in some area that contains vegetation far from the scan center, or when buildings are positioned behind other objects,
the limited number of scan points within the bin leads to potential inaccuracies in determining the status of the bins.
% In some area that contains vegetation far from the scan center, or when buildings are positioned behind other objects,
% the limited number of scan points within the bin leads to potential inaccuracies in determining the status of the bins.
% the number of scan point in the query bin is pretty low and the status of bins may be wrong,
% which results in mistaken removal of static object points.
}
% \item{{\bf{Uncertainty and erroneous removal in areas with unstructured objects.}} As shown in Fig.\ref{0limitation}(a), in some area that contains vegetation, especially when the query bin is far from the scan center, the limited number of scan points within the bin leads to potential inaccuracies in determining the status of the bins,
% % the number of scan point in the query bin is pretty low and the status of bins may be wrong,
% which results in mistaken removal of static object points.
% }
\item{{\bf{Isolated dynamic bin caused by drift or error of LiDAR data.}} As shown in Fig.\ref{3Around}, certain bins may exhibit an isolated dynamic status despite the absence of any dynamic objects in that particular area, 
% there may be some isolated bins in dynamic status with no dynamic object in that area,
which leads to wasted computation and ineffective removal processes.
% which may result in a waste of computation and bad removal. 
}
% \item{{\bf{Wrong judgment when lower part is blocked.}} As shown in Fig.\ref{0limitation}(c), particularly when buildings are positioned behind other objects, the wall of buildings may easily be mistakenly identified as dynamic objects and subsequently removed, which decreases the number of preserved static points.
% % there is a risk of erroneously identifying the walls of the buildings as dynamic objects, leading to their removal.
% % especially when buildings are behind some objects, the wall of buildings may easily be regarded as dynamic objects and removed.
% }
\end{itemize}

% \begin{figure*}[!t]
% \centering
% \includegraphics[width=7in]{0_v2-3Limitation.png}
% \caption{Visualization of Limitation in ERASOR. (a) The uncertainty in areas with unstructured objects. (b) The isolated dynamic bin caused by drift or error of LiDAR data. (c) The wrong judgment when the lower part is blocked. (d) Good removal on the ground.}
% \label{0limitation}
% \end{figure*}

To deal with these limitations above, we modified the previous method and proposed ERASOR++, an improvement based on a height coding method.

\subsection{LiDAR Descriptors}
% LiDAR descriptors have proven to be extremely valuable in representing vast amounts of information within 3D point clouds. They find applications in various domains, including location estimation and registration. Broadly speaking, there are two primary categories of LiDAR descriptors: hand-crafted based methods and deep-learning based methods.
% In the context of hand-crafted based approaches, the classification of descriptors can be performed based on the region from which information is extracted. This classification typically includes local descriptors, global descriptors, and hybrid descriptors.
% Local descriptors focus on capturing fine-grained details within localized regions of the point cloud. These descriptors excel at providing accurate representations of specific objects or areas of interest. Global descriptors, on the other hand, aim to capture the overall structure and characteristics of the entire point cloud. They provide a holistic view that can be useful for tasks such as scene classification or large-scale mapping. Hybrid descriptors combine the advantages of both local and global descriptors, aiming to strike a balance between capturing detailed local information and overall global structure.
% By classifying descriptors based on the extracted region, these hand-crafted based approaches offer flexibility and adaptability to various requirements in different 3D applications and scenarios.
LiDAR descriptors have been a considerably vital method to represent numerous 3D point cloud information, used in various 3D applications such as location and registration, which can be divided into two key categories, namely hand-crafted and deep-learning based methods.
Focused on the traditional hand-crafted approaches, classification can be made by taking into account the extracted region, such as local, global, and hybrid descriptor \cite{reviewdes}.

% Local descriptors are always designed to encode the local information of feature points such as surface normal and curvature, generating based on spatial distribution information, geometry information as well as intensity information \cite{risas}. And the familiar Point Feature Histogram (PFH) \cite{pfh} \cite{pfh2} and Fast Point Feature Histogram (FPFH) \cite{fpfh} methods are classical geometry based 3D point descriptors.
% % Caused by searching and estimating their neighborhood points, local descriptors always need large computational quantity. 
% Global descriptors encode the information of the whole 3D point cloud and require less computation time, which can also include three types of information above, such as Viewpoint Feature Histogram (VFH) \cite{vfh} method based on geometry and Global Structure Histogram (GSH) \cite{gsh} method based on spatial distribution. 
% These global methods are increasingly used in 3D recognition, geometric categorization, and shape retrieval \cite{reviewdes}. 
% Hybrid descriptors combine the advantages of both local and global descriptors, aiming to strike a balance between capturing detailed local information and overall global structure.

Moreover, there are some global descriptors designed for specific applications, like Scan Context \cite{sc}  and its expansion method \cite{iris} \cite{isc}, which are primarily designed for loop detection tasks.
These descriptors leverage spatial distribution or intensity information within each point block, serving as a source of inspiration for the presentation and description of LiDAR point clouds in this paper.
%And there are some global descriptors designed for specific applications, like Scan Context \cite{sc}  and its expansion method \cite{iris} \cite{isc} designed for loop detection tasks, leveraging spatial distribution or intensity information in each point block, which inspire the presentation and description of LiDAR point cloud in this paper.

\section{Methodology in ERASOR++}\label{Methodology}
% In this section, the problem of dynamic object removal in point cloud map is described and methodology in ERASOR++  are represented in detail in the following 4 parts.
In this section, the methodology in ERASOR++ is represented in detail in the following parts.
% \subsection{Methodology Overview}
The framework of our ERASOR++ method, as well as the connections between methods and limitations in Section \ref{limitationslist} are depicted in Fig.\ref{0Framework}. 
% The framework of our ERASOR++ method is depicted in Fig.\ref{0Framework}. And the connections between methods and limitations in Section \ref{limitationslist} are illustrated in Fig.\ref{0Framework}. 
The subsequent analysis reveals that each method, addressing a specific limitation, contributes to the optimization of the dynamic object removal system from different perspectives.
% It is shown that each method related to one limitation could optimize dynamic object removal system in proper aspect.

\subsection{Height Coding Descriptor}\label{HCD}
Raw point clouds obtained from scans and prior maps often contain an overwhelming amount of structural information, making calculations complex and challenging. Hence, it is necessary to simplify the information and adopt a novel method to describe the point cloud in the area of interest. Motivated by \cite{erasor} \cite{sc} \cite{iris}, we propose a novel approach by integrating the height difference and height encoding information, resulting in the Height Coding Descriptor (HCD) which offers a both concise and informative representation of the point cloud in the region of focus.
% As raw point clouds from scans and prior maps contain too much structure information which may be difficult and complicated when calculating, it is necessary to simplify the information and use a novel method to describe the point cloud in interest area. Inspired by \cite{erasor} \cite{sc} \cite{iris}, we combine the height difference with height encoding information and make Height Coding Descriptor (HCD).

\begin{figure}[!t]
\centering
\includegraphics[width=3.5in]{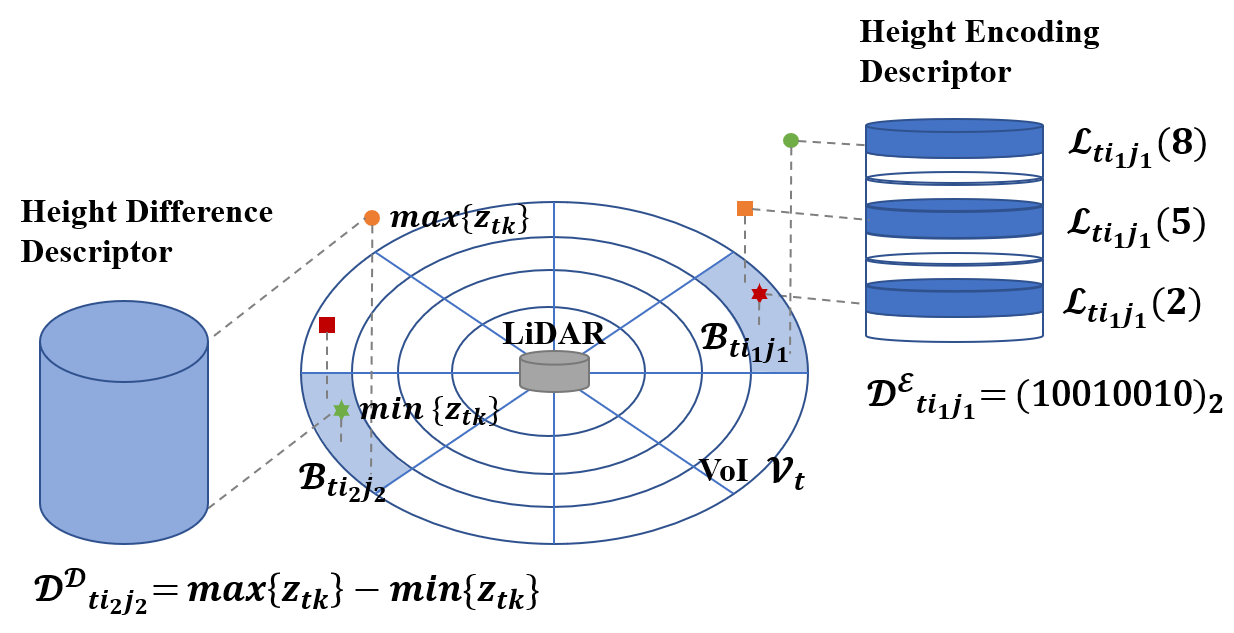}
\caption{The representation of VoI range and Bin division of point cloud, along with the definition of combined Height Difference Descriptor and Height Encoding Descriptor, as well as a numeral example to clarify encoding part.}
\label{1Descriptor}
\end{figure}

When the current scan comes, we first fetch the point cloud within a distance range, then get the submap from the prior map by fetching the point cloud in the same range. The Height Coding Descriptor is based on the fetched point cloud as presented in Fig.\ref{1Descriptor}.

Let $\mathcal{P}_t$ be the point cloud perceived at time step $t$; ${p_t}_k$ be a point in $\mathcal{P}_t$, i.e., ${p_t}_k \in \mathcal{P}_t$; and $\mathcal{V}_t$ be the points fetched from a specific range, also called Volume of Interest (VoI). 
Then, $\mathcal{V}_t$ can be defined and formulated as:
\begin{equation} %leave a line or not?
\begin{aligned}
\label{voi}
\mathcal{V}_t = \{{p_t}_k | {p_t}_k\in\mathcal{P}_t, {\rho_t}_k<L_{max},\\
H_{min}<{z_t}_k<H_{max}\},
\end{aligned}
\end{equation}
where ${p_t}_k = \{ {x_t}_k, {y_t}_k, {z_t}_k\}$, and ${\rho_t}_k=\sqrt{{{x_t}_k}^2+{{y_t}_k}^2}$. Note that $L_{max}$, $H_{min}$ and $H_{max}$ are constant threshold of VoI.
This formula limits the maximal radial distance boundary using $L_{max}$ and the valid height range of VoI using $H_{min}$ and $H_{max}$, which can cut down the number of points to be transformed and avoid the outlier.

As shown in Fig.\ref{1Descriptor}, the VoI $\mathcal{V}_t$ will be divided into bins according to azimuthal and radial position, known as sectors and rings. Let ${\mathcal{B}_t}_{ij}$ be the points of bins in $i$ ring and $j$ sector, then the points in each bin can be defined as:
\begin{equation}
\begin{aligned}
\label{bin}
{\mathcal{B}_t}_{ij} = \{{p_t}_k | {p_t}_k\in\mathcal{V}_t, \frac{(i-1)L_{max}}{N_r}\leq{\rho_t}_k<\frac{i L_{max}}{N_r}, \\
\frac{(j-1) 2\pi} {N_\theta}\leq{\theta_t}_k<\frac{j 2\pi} {N_\theta}\},
\end{aligned}
\end{equation}
where ${\theta_t}_k = arctan2( {y_t}_k,{x_t}_k)+\pi$, $N_r$ and $N_\theta$ respectively mean the total divisional numbers of rings and sectors.

In each bin, layer indexes based on height value are utilized to represent the height information in the descriptor. Let ${\mathcal{L}_t}_{ij}(\alpha)$ be the ${\mathcal{B}_t}_{ij}$ points in layer index $\alpha$ and designed as:
\begin{equation}
\begin{aligned}
\label{layer}
{\mathcal{L}_t}_{ij}(\alpha) = \{{p_t}_k | {p_t}_k\in{\mathcal{B}_t}_{ij}, 
\frac{(\alpha-1)(H_{max}-H_{min})}{N_{l}} \\
\leq{z_t}_k < \frac{\alpha (H_{max}-H_{min})}{N_{l}}\}, 
\end{aligned}
\end{equation}
where $N_{l}$ means the total layer number of indexes in each bin.

So the value of Height Coding Descriptor $\mathcal{D}_t$ as well as its difference descriptor ${{\mathcal{D}^\mathcal{D}}_t}_{ij}$ and encoding descriptor ${{\mathcal{D}^\mathcal{E}}_t}_{ij}$ of the bins in $i$ ring and $j$ sector can be defined as:
\begin{align}
\label{des}
{{\mathcal{D}^\mathcal{D}}_t}_{ij} &= max\{{z_t}_k\}-min\{{z_t}_k\},\ {p_t}_k\in{\mathcal{B}_t}_{ij} \\
{{\mathcal{D}^\mathcal{E}}_t}_{ij} &= \sum_{\alpha=1}^{N_{l}} {\mathcal{O}_t}_{ij}(\alpha) \cdot 2^{\alpha-1},\\ 
{\mathcal{O}_t}_{ij}(\alpha) &= \begin{cases}
1,&{\text{if}} \ \exists \ {p_t}_k\in{\mathcal{L}_t}_{ij}(\alpha) \\ 
{0,}&{\text{otherwise.}} 
\end{cases},\\
\mathcal{D}_t &= \{ {{\mathcal{D}^\mathcal{D}}_t}_{ij}, {{\mathcal{D}^\mathcal{E}}_t}_{ij}\}.
\end{align}
% Note that ${p_t}_k\in {\mathcal{B}_t}_{ij}$, and $Layer(\alpha)$ means the height layer space in each bin measured in $Z$ axis, as depicted in Fig.\ref{1Descriptor}. 
Note that the representation of ${{\mathcal{D}^\mathcal{E}}_t}_{ij}$ is defined for using bit operation to simplify computation and save storage.
As shown in Fig.\ref{1Descriptor}, the height occupancy information in each layer can be represented completely using this encoding method with an optional number of total layers $N_{l}$, which is always set as 8 or its multiples to match computer storage unit. 

Thus, the descriptor can not only show the extreme value of relative height but also represent other points' height information with a simplified encoding method.

\subsection{Height Stack Test}\label{HST}
\begin{figure}[!t]
\centering
\includegraphics[width=3.3in]{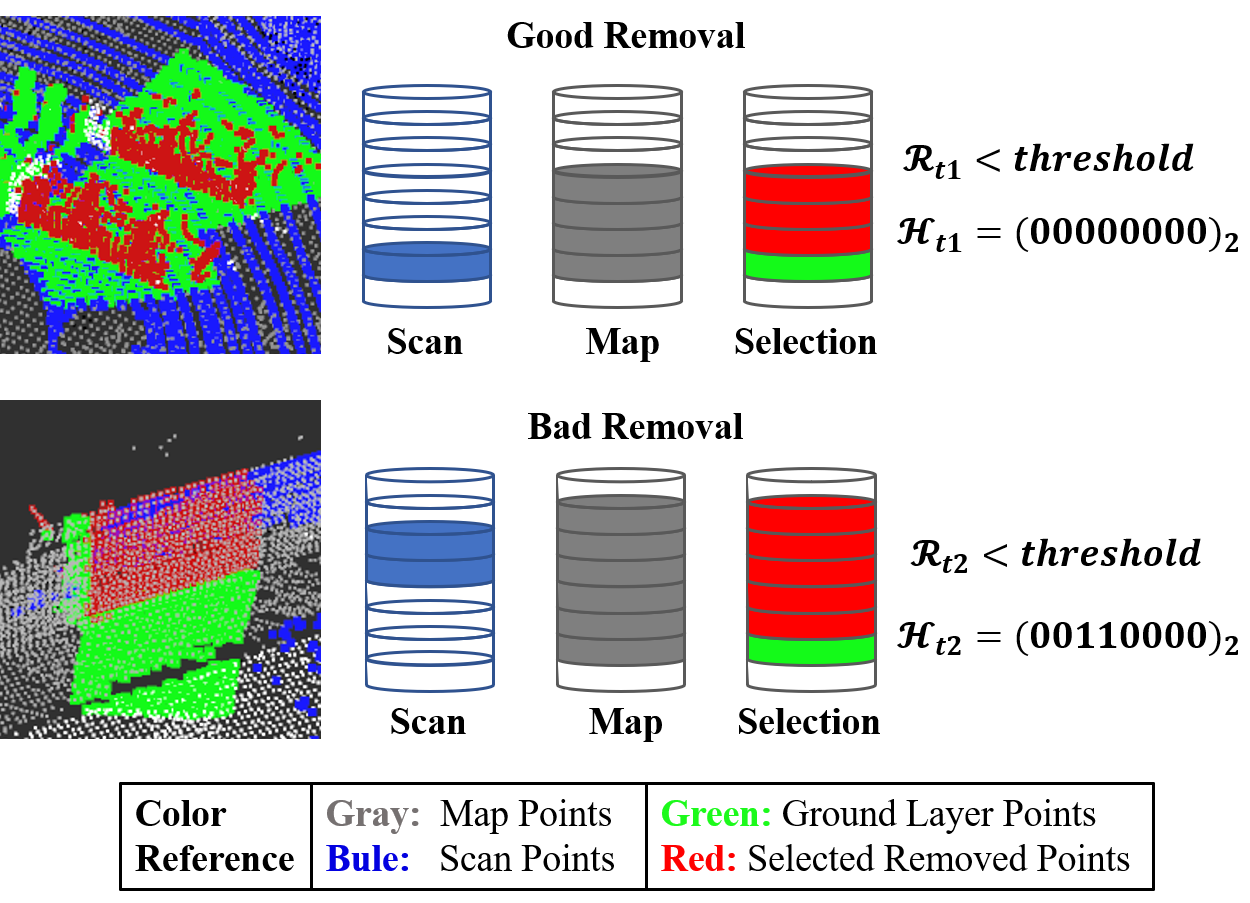}
\caption{The good removal and bad removal, which can not be distinguished through Scan Ratio Test, but can be greatly separated through Height Stack Test, according to the formula and example shown in this figure. Note that $threshold$ is a constant value and empirically set to 0.2 in \cite{erasor}. When the overlap layer is only around the ground, the dynamic removal is good and valid. On the contrary, when the overlap is in high layers, there may be bad removal.}
\label{2StackRatio}
\end{figure}

In previous work, Scan Ratio Test (SRT) was used to compare the current scan and its corresponding prior map descriptors, in order to found out bins with dynamic objects. SRT directly compares the values of height difference in the current scan and map scan by restricting the quotient of these two values, which can be described as:
\begin{equation} 
\begin{aligned}
\label{scanratio}
\mathcal{R}_t = \frac{{\mathcal{D}^\mathcal{D}}_t(Curr)}{{\mathcal{D}^\mathcal{D}}_t(Map)},
\end{aligned}
\end{equation}
which may cause bad removal as displayed in Fig.\ref{2StackRatio}. In that case, based on the proposed HCD, a novel test method called Height Stack Test (HST) is designed and utilized.

From Fig.\ref{2StackRatio}, we can see that the overlaps in layers above ground need to be checked, and the quantity of overlap parts demonstrates the possibility of being static. So the Height Stack Test parameter is designed as:
\begin{equation} 
\begin{aligned}
\label{stackratio}
% \mathcal{H}_t = \neg [ {\mathcal{D}^\mathcal{E}}_t(Curr)\oplus{\mathcal{D}^\mathcal{E}}_t(Map) ]\\
\mathcal{H}_t = [ {\mathcal{D}^\mathcal{E}}_t(Curr)\wedge{\mathcal{D}^\mathcal{E}}_t(Map) ]\\
\wedge(\neg{Layer(Ground)}),
\end{aligned}
\end{equation}
where the sign $\wedge$ and $\neg$ respectively means the AND and NOT operation of bit coding, and $Layer(Ground)$ means the encoding layer with point cloud below the ground, which is described in the following part \ref{GLT}. 
$\mathcal{H}_t$ represent overlap layers as well as their position above ground.
% , more overlap layers demonstrate greater tendency for bins to be static. 
Note that $\mathcal{H}_t$ is always turned to its number of occupied bits to be checked and limited to a low level when coding and implementing.

Thus, with restriction of the position and number of overlap layers, the height information in the middle position can be leveraged, and bad removal caused by blocked points in scan bins can be successfully avoided.
% overlaps in other layers except for the ground layer can be successfully avoided.

\subsection{Ground Layer Test}\label{GLT}
Ground Layer Test (GLT) is a proposed method that involves examining the ground layer within the HCD to determine the relative position for the HST. 
% Ground Points Test is a proposed method according to the height encoding descriptor, in order to find the descriptor layer with ground points. 
As the Z direction information of pose from LiDAR Odometry or SLAM is usually more uncertain than X and Y direction \cite{loam}, the ground layer in each sequence is not united, which causes the necessity of searching ground layer in this system.

From the process above, each point in VoI is related to one bin and can be indexed from the ring number $i$ and sector number $j$ in ${\mathcal{B}_t}_{ij}$. Considering that ground points can always be fetched near the scan sensors, we use the index of each bin to iterate from near to far and get the occupied ring $i$, then check the number of points in each layer of these bins, and finally select the layer with enough most points as the ground layer in ${\mathcal{B}_t}_{ij}$.
If more than three-quarters of bins in $i$ ring achieve the same ground layer $\gamma$, the value of $Layer(Ground)$, which encodes layers below the ground, is defined as:
\begin{equation} 
\begin{aligned}
\label{groundlayer}
Layer(Ground) &= \sum_{\alpha=1}^{\gamma} 2^{\gamma-1},
\end{aligned}
\end{equation}
where $\gamma$ is the sequence of layers with the satisfied number of ground points in the ring of nearest valid bins. 

Thus, the ground layer is found and then the height descriptor can represent both absolute and relative information. Note that different from Ground Plane Fitting (R-GPF) in \cite{erasor}, our Ground Layer Test method only leverages a simple calculation to find the layer with ground points and runs independently before all the processes to give restrictive information for the following HST and R-GPF.

\subsection{Surrounding Points Test} \label{SPT}
\begin{figure}[!t]
\centering
\includegraphics[width=3.0in]{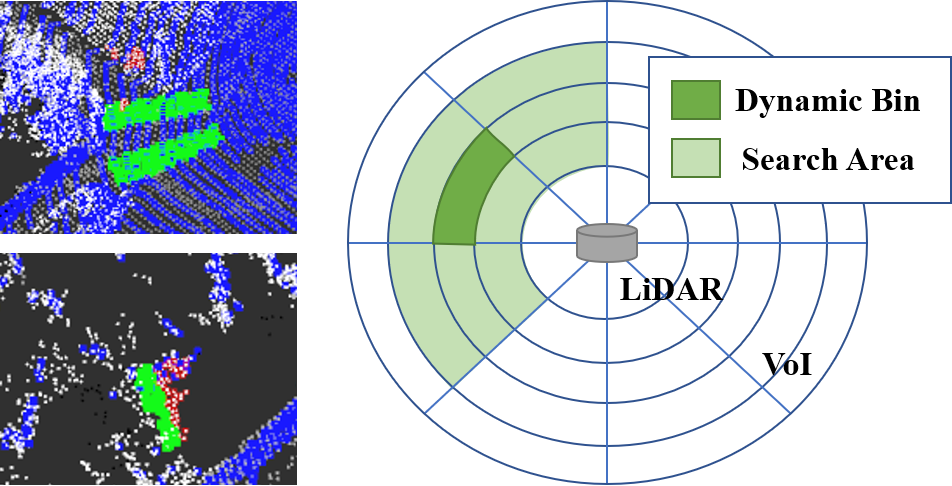}
\caption{(Left) Isolated dynamic bin in point cloud, which may cause bad removal. (Right) The searching area of each dynamic bin.}
\label{3Around}
\end{figure}

Surrounding Points Test (SPT) is a process between HST and R-GPF to cut down the wrong dynamic removal. 
As uncertainty always exists in LiDAR point cloud and some shelters may affect the result of bin status, there are some isolated bins in dynamic status but with no dynamic object in it, as shown in Fig.\ref{3Around}.

This part focuses on correcting isolated bin status by searching for neighboring points, aiming to identify and rectify situations where a bin appears to be dynamic due to insufficient surrounding information. By analyzing points in the vicinity, this test ensures a more accurate assessment of the bin's dynamic status.

For the prior map is generated by the accumulation of each scan frame, dynamic objects always have long tracks in maps. So the bins indeed with dynamic objects should be congregated as groups or lines. Under this circumstance, we proposed the Surrounding Points Test method to check each dynamic bin and make sure their status is right by searching their surrounding points. The search area can be described as:
\begin{equation} 
\begin{aligned}
\label{surround}
{\mathcal{S}_t}_{ij} = \{{\mathcal{B}_t}_{pq} | (i-Range)\leq p < (i+Range),\\
(j-Range)\leq q < (j+Range)\},
\end{aligned}
\end{equation}
where $Range$ means the point cloud searching range which can be changed according to the data sequence, and $Range=1$ is enough for most situations.

If there are dynamic bins in searching areas, the status of the target bin will remain unchanged. On the contrary, if there are no dynamic bins in searching areas, the status of the target bin will be changed to static status.

Thus, the isolated wrong removal is decreased and more static points are reserved in the output map with an increase in preservation rate.

\section{Experiments and Discussion}\label{Experiments}
% In this section,  we compare our proposed dynamic removal system with the initial ERASOR and evaluate the algorithm through ablation experiments.
In this section,  we evaluate our proposed dynamic object removal system through comparison and ablation experiments.
All the experiments were conducted on a PC with 2.2GHz cores, 16GB RAM, and Robot Operating System (ROS) \cite{ros} running in Ubuntu 18.04. The following parts present details of all the experiments.

\subsection{Settings and Datasets}
Two parts of experiments were set to evaluate and compare the proposed algorithm, including overall system tests and ablation experiments. 
% using datasets tested in previous work and adding other datasets untested before. 
Overall system tests evaluated the precision and speed of the system with all methods proposed above, while ablation experiments evaluated the effect of partial methods. The initial ERASOR \cite{erasor} was tested in all datasets for comparison.

According to the system framework in Fig.\ref{0Framework}, the HCD and GLT parts are preconditions of HST, and SPT can be solely added in ERASOR \cite{erasor}. 
So the ablation experiments include the evaluation of system without HCD, GLT, and HST (in Section \ref{HCD}, \ref{GLT} and \ref{HST}) and system without SPT (in Section \ref{SPT}). Note that they were respectively named as ++ w/o ABC and ++ w/o D in Table \ref{tab:table1}.

The SemanticKITTI dataset \cite{kitti} \cite{kitti2012} was chosen for experiments, as it provides annotations for each point and labels the ground-truth dynamic objects that need to be removed, which is feasible for us to evaluate the algorithm system.
Similar to the datasets used in ERASOR \cite{erasor}, we used the selected frames with the largest part of dynamic object in different sequences, that is, 
4390-4530 frames in sequence 00, 150-250 frames in sequence 01, 860-950 frames in sequence 02, 2350-2670 frames in sequence 05, and 630-820 frames in sequence 07. 

% Different from the previous work, to evaluate the generalization of the proposed method, all sequences with ground truth are tested in our experiments.
% For the sequences used in \cite{erasor}, similarly, the selected frames with the largest part of dynamic objects in different sequences were utilized, that is, 
% 4390-4530 frames in sequence 00, 150-250 frames in sequence 01, 860-950 frames in sequence 02, 2350-2670 frames in sequence 05, and 630-820 frames in sequence 07. 
% For other sequences, we completed the KITTI sequences with ground truth from 00 to 10, by adding other 6 sequences. The selected segments with dynamic objects include 30-130 frames in sequence 03, 170-270 frames in sequence 04, 840-980 frames in sequence 06, 0-160 frames in sequence 08, 1350-1480 frames in sequence 09,  and 910-1100 frames in sequence 10.

Note that the experiments were conducted in these frames above but recorded directly using the number of sequences in Table \ref{tab:table1}.
% With reference to the previous work, prior maps were conducted and accumulated using the poses calculated by SuMa \cite{suma}. 
% For each sequence, proper adjustment on added parameters was checked and utilized. Parameters in previous work \cite{erasor} except for those added in the proposed algorithm, remained unchanged for fair comparison.
For a fair comparison, parameters in previous work \cite{erasor} remained unchanged.

\subsection{Algorithm Precision}
Algorithm precision was measured through the novel static map-oriented quantitative metrics in \cite{erasor} called Preservation Rate (PR) and Rejection Rate (RR). 

% \usepackage{booktabs}
% \usepackage{multirow}
% \usepackage{booktabs}
% \usepackage{multirow}
\begin{table}[!t]
\caption{Experiments Results and Comparison in KITTI seq.\label{tab:table1}}
\centering
\begin{tabular}{@{}cccccc@{}}
\toprule
Seq.                & Method      & PR{[}\%{]}       & RR{[}\%{]}       & F1 score          & Avg. Time{[}s{]} \\ \midrule
\multirow{4}{*}{00} & ERASOR     & 92.1498          & 97.206  & 0.946104          & 0.125032        \\
                    & ERASOR++   & \textbf{96.8261} & 96.1009 & \textbf{0.964621} & 0.124875        \\
                    & ++ w/o ABC & 95.8274          & 96.0592 & 0.959432          & 0.115372        \\
                    & ++ w/o D   & 93.5736          & 96.789  & 0.951542          & 0.135319        \\ \midrule
\multirow{4}{*}{01} & ERASOR     & 91.8967          & 94.5626 & 0.932106          & 0.132209        \\
                    & ERASOR++   & \textbf{98.9919} & 93.6401 & \textbf{0.962471} & 0.13711         \\
                    & ++ w/o ABC & 97.3527          & 93.8679 & 0.955785          & 0.126481        \\
                    & ++ w/o D   & 94.9805          & 94.139  & 0.945579          & 0.143876        \\ \midrule
\multirow{4}{*}{02} & ERASOR     & 80.896           & 99.2045 & 0.891197          & 0.161044        \\
                    & ERASOR++   & \textbf{87.8949} & 98.9015 & \textbf{0.930739} & 0.135772        \\
                    & ++ w/o ABC & 82.6932          & 99.2045 & 0.901995          & 0.154411        \\
                    & ++ w/o D   & 85.9189          & 98.9015 & 0.919542          & 0.13935         \\ \midrule
\multirow{4}{*}{05} & ERASOR     & 86.9621          & 97.9208 & 0.921167          & 0.121533        \\
                    & ERASOR++   & \textbf{96.5283} & 97.6666 & \textbf{0.970941} & 0.10012         \\
                    & ++ w/o ABC & 91.2505          & 97.7049 & 0.943673          & 0.113161        \\
                    & ++ w/o D   & 94.4625          & 98.1019 & 0.962478          & 0.10433         \\ \midrule
\multirow{4}{*}{07} & ERASOR     & 93.4848          & 98.8887 & 0.961109          & 0.090536        \\
                    & ERASOR++   & \textbf{98.5774} & 98.6509 & \textbf{0.986141} & 0.101062        \\
                    & ++ w/o ABC & 96.6388          & 98.768  & 0.976918          & 0.088366        \\
                    & ++ w/o D   & 97.583           & 98.6935 & 0.981351          & 0.105337        \\ \bottomrule
\end{tabular}
\end{table}

\begin{figure*}[!t]
\centering
\includegraphics[width=7in]{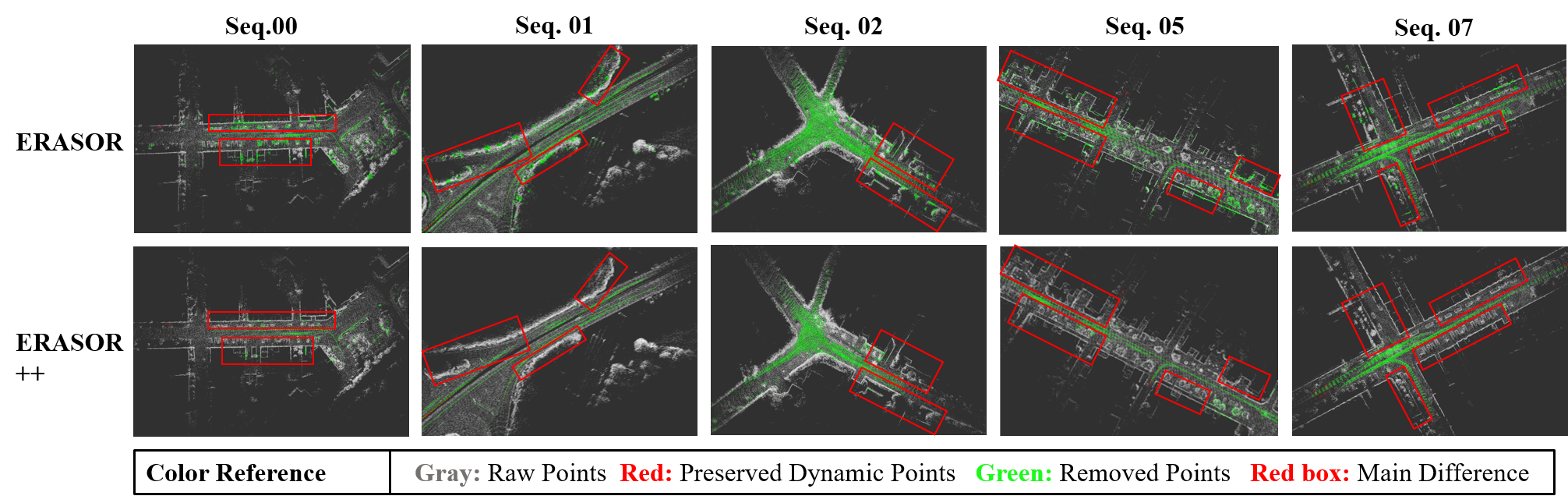}
\caption{Visual Comparison of Experiments Results using ERASOR and ERASOR++. Through the main difference in red boxes, our proposed method can largely preserve static structures, such as buildings and vegetation.}
\label{6Visual}
\end{figure*}

\begin{figure}[!t]
\centering
\includegraphics[width=3.5in]{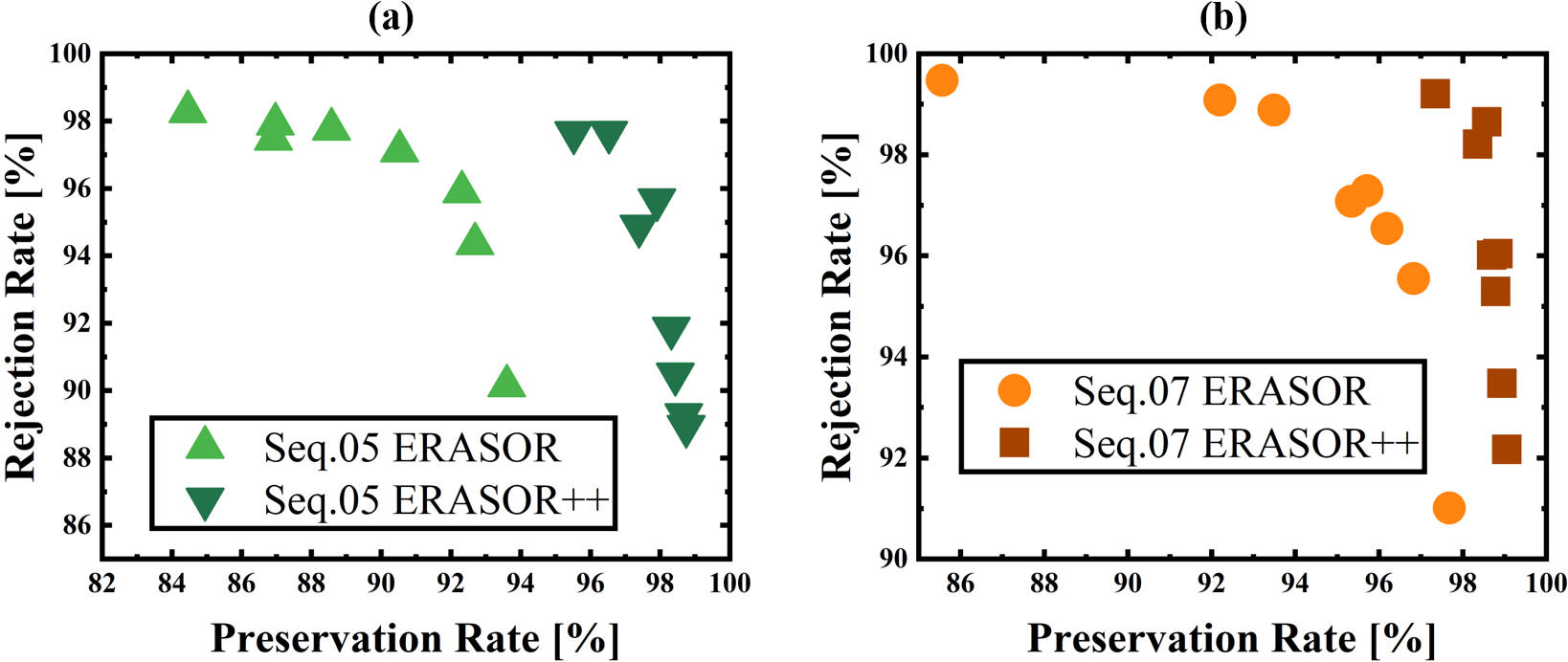}
\caption{General Comparison of Experiment Results using ERASOR and ERASOR++. The PR and RR results of sequence 05 and sequence 07 with different external interval parameters are depicted. The results of ERASOR++ always display on the high-value side, which shows that ERASOR++ wins better precision in general.}
\label{5-2Results}
\end{figure}

PR describes the effect of static point cloud preservation as an evaluation of bad removal, which is the ratio of the preserved static points in all static points. RR describes the effect of dynamic point cloud removal, which is the ratio of removed dynamic points in all dynamic points. 
Both of these metrics were calculated voxel-wise with a voxel size of 0.2.
$F_1$ score is the combination metric of PR and RR.
% These metrics can be indicated as:
% \begin{align}
% \label{metrics}
% \text{PR} &= \frac{Size(\mathcal{P}_t^{ps})}{Size(\mathcal{P}_t^{as})},\\
% \text{RR} &=1- \frac{Size(\mathcal{P}_t^{pd})}{Size(\mathcal{P}_t^{ad})},\\
% F_1 &= 2 \cdot \frac{\text{PR}\cdot\text{RR}}{\text{PR} + \text{RR}},
% \end{align}
% where $\mathcal{P}_t^{ps}$ means preserved static points, $\mathcal{P}_t^{as}$ means all static points in raw map, $\mathcal{P}_t^{pd}$ means preserved dynamic points and $\mathcal{P}_t^{ad}$ means all dynamic points in raw map.

The results of the previous and proposed algorithm as well as the ablation experiments are listed in Table \ref{tab:table1},
in which the previous algorithm was replicated using its open source code.
% which provide a comprehensive comparison of the performance of both methods using concrete data in detail.
% the ground truth static and dynamic points numbers, as well as preserved static and dynamic points numbers after using dynamic removal algorithm are depicted in Fig.\ref{5Results}, 
The visual comparison of results with main differences in red boxes is depicted in Fig.\ref{6Visual},
% which clearly represents the algorithm's effectiveness.
which provides a clear representation of the effectiveness of the algorithm.
%Note that the results on our experiments platform were a little bit different from that in paper \cite{erasor}.
% It is shown that in sequences 00, 01, 03, 04, 08, and 10 our method consistently achieves better results across all three metrics, as well as in other sequences wins better results in combination $F_1$ score.
It is shown that in all sequences, our method consistently achieves extreme improvements in Preservation Rate and F1 score, with only a slight decrease in RR value.
% which can be clearly represented in the line chart in Fig.\ref{5Results}. 
And Preservation Rate in our method can always be kept at a high level. Although in sequence 02, there is a defective PR caused by the pose uncertainty in the Z axis,  the results still greatly outperform the previous work.
Moreover, considering the difference in points quantity, the improvement in the preserved static points number by our algorithm far outweighs the changes in the dynamic points number.

Note that when testing in long-term datasets, there is no need to run the dynamic removal system in each frame. As an external parameter, the frame skip interval could also influence the PR and RR values of each method. We tested and compared these two methods with different interval parameters, and the results of sequences 05 and 07 are depicted in Fig.\ref{5-2Results}. It shows that through the trend of PR-RR relationship, the results of ERASOR++ always display on the high-value side and win high-level precision compared with ERASOR. 
% Additionally, even in sequences 02, 05, 06, 07, and 09 where our method has a slightly lower RR value compared to the previous work, the RR value is still maintained at a promising high level while achieving a better $F_1$ score, which can be clearly represented in the line chart in Fig.\ref{5Results}. 
% Note that although in sequences 02, 05, 06, 07, and 09 our method gets a little bit lower RR value than previous work, the RR value is still kept at a promising high level with a better $F_1$ score, which can be represented clearly from line chart in Fig.\ref{5Results}. And the points numbers depicted in Fig.\ref{5Results} show that  the improvement of preserved static points number is far beyond the change of dynamic points number.
% Moreover, the points numbers shown in Fig.\ref{5Results} illustrate that the improvement in the preserved static points number by our algorithm far outweighs the changes in the dynamic points number.

That is, our proposed algorithm demonstrates superior precision compared to the previous ERASOR \cite{erasor} algorithm across various sequences.
% That is, the precision of our proposed algorithm performs better than that of the previous ERASOR \cite{erasor} algorithm.

\subsection{Algorithm Speed}

\begin{table}[!t]
\caption{Time cost and correlative factor R-DPF count.\label{tab:table2}}
\centering
\begin{tabular}{@{}ccccc@{}}
\toprule
Method          & \multicolumn{2}{c}{Seq. 05}               & \multicolumn{2}{c}{Seq.07}                \\ \midrule
Average Eva.  & R-GPF Count & Time{[}s{]} & R-GPF Count & Time{[}s{]} \\ \midrule
ERASOR          & 577.9              & 0.121533            & 531.632             & \textbf{0.090536}   \\
ERASOR++        & \textbf{368.875}    & \textbf{0.10012}   & \textbf{390.053}    & 0.101062            \\ \bottomrule
\end{tabular}
\end{table}

Algorithm speed was measured through the average time taken for one frame iteration during the execution of the main algorithm. 
% costing in one frame iteration when running the main algorithm part. 
We also compared the speed of our proposed method with the previous ERASOR work and other ablation systems, as shown in Table \ref{tab:table1}. 
% It should be explained that the average time of these two algorithms was calculated on the same experimental platform, using the same recording method implemented by our team. 
% Therefore, it is reasonable to expect some slight differences between the data reported in \cite{erasor} and our results, which proves the validity of the comparison.

% However, the comparison still holds validity.
% of our experiment with the same recording method compiled by us, so it is reasonable that there may be a little bit of difference between the data in \cite{erasor}.
% Therefore, it can prove that our algorithm costs equivalent time or even less time than the previous work. 

On account of the main time-costing calculation part R-GPF, the count of this part decides the time cost of algorithms.
As shown in Table \ref{tab:table2}, although our proposed methods have some incremental calculations compared with the previous method, our algorithm demonstrates an equivalent or even faster execution time compared to the previous ERASOR algorithm, for the iteration counts of R-GPF are less than ERASOR methods. 
% Note that the relationship between the R-GPF count and time cost is not linear, for there are other factors concerning time cost.
% such as the quantity of points and the range of maps.
Considering that the previous ERASOR algorithm is already known to be at least ten times faster than other methods \cite{erasor}, our proposed algorithm, ERASOR++, also achieves a comparable level of efficiency.
% For the previous ERASOR is at least ten times faster than other methods \cite{erasor}, our proposed ERASOR++ also gains comparative fast performance.

This indicates that our algorithm not only improves upon the previous algorithm in terms of accuracy and precision but also maintains a comparable level of speed performance.

% \begin{figure}[!t]
% \centering
% \includegraphics[width=3.5in]{0914_5Result.png}
% \caption{Statistical Comparison of Experiments Results using ERASOR and ERASOR++. The Preservation Rate outperformance and the Rejection Rate maintenance can be shown directly and clearly.}
% \label{5Results}
% \end{figure}

\subsection{Ablation Experiments}
% Ablation experiments include ERASOR++ system without HCD, GLT and HST (in Section \ref{HCD}, \ref{GLT} and \ref{HST}, named as ++ w/o ABC) and system without SPT (in Section \ref{SPT}, named as ++ w/o D).
% The results of ablation experiments are listed in Table \ref{tab:table1}.
% Ablation experiments include ERASOR++ system without HCD, GLT and HST (in Section \ref{HCD}, \ref{GLT} and \ref{HST}, named as ++ w/o ABC) and system without SPT (in Section \ref{SPT}, named as ++ w/o D).
The results of ablation experiments are listed in Table \ref{tab:table1}.
From all the experiment sequences, the results of our system without each part demonstrate a decrease in Preservation Rate, which means these parts can avoid different kinds of bad removal and solve corresponding limitations. 
% And the results of system without HCD and relative parts perform comparative results in some sequence such as 04, 05, 06, and 07, but demonstrate a decrease of Reject Rate in other sequence. That is, our total system can better adapt in various kinds of environment, keeping high level precision in all test sequences.

Through these ablation experiments, the validation of each part can be proved.
The Surrounding Points Test and Height Coding Descriptor as well as its corresponding test methods effectively decrease the bad removal and improve the system generalization. 

\section{Conclusions}\label{Conclusion}
In this work, an improved dynamic object removal system ERASOR++ was proposed. Building upon the previous work of ERASOR \cite{erasor}, our method introduced a novel representation called Height Coding Descriptor, along with the comprehensive Height Stack Test, Ground Layer Test, and Surrounding Points Test. The proposed system addressed the limitations of ERASOR and effectively promoted the quality of preserved static points with total structure, particularly in unstructured environments, which can greatly contribute to map reconstruction. This improvement offers a more accurate and robust alternative for dynamic object removal when constructing static maps.
% based on the previous work \cite{erasor} with a novel representation of Height Coding Descriptor, a comprehensive Height Stack Test method for dynamic bins, additional Ground Points Test and Surrounding Points Test for accurate results. 

% Experiments were conducted using the open-source SemanticKITTI dataset, and 
% The experiments results demonstrated the superiority of our proposed system compared to the previous work. This improvement offers a more accurate and robust alternative for dynamic object removal when constructing static maps.
% Our system could deal with limitations of \cite{erasor}, like bad removal in unstructured environments. It was tested using open-source SemanticKITTI dataset and won better results than the previous work, which offers a more accurate and robust alternative to dynamic object removal when building static maps.

In future works, with the innovation of novel representation, the descriptor is supposed to be completed through more available information from raw point clouds, and further, be utilized in other challenging tasks or aspects, providing further opportunities for exploration and utilization.

% In this study, we presented an enhanced dynamic object removal system called ERASOR++. Building upon the previous work of ERASOR \cite{erasor}, our method introduced a novel representation called Height Coding Descriptor. This descriptor, along with the comprehensive Height Stack Test, Ground Points Test, and Surrounding Points Test, addressed the limitations of ERASOR, particularly in unstructured environments.

% We conducted experiments using the open-source SemanticKITTI dataset, and the results demonstrated the superiority of our proposed system compared to the previous work. This improvement offers a more accurate and robust alternative for dynamic object removal when constructing static maps.

% In future research, we plan to continue innovating the representation of the descriptor. By incorporating more available information from raw point clouds, we aim to enhance its completeness. Additionally, we believe that this improved descriptor can be applied to other challenging tasks or aspects, providing further opportunities for exploration and utilization.

\section*{Acknowledgments}
This work was supported by STI 2030-Major Projects 2021ZD0201403, in part by NSFC 62088101 Autonomous Intelligent Unmanned Systems.
% This should be a simple paragraph before the References to thank those individuals and institutions who have supported your work on this article.

% {\appendix[Proof of the Zonklar Equations]
% Use $\backslash${\tt{appendix}} if you have a single appendix:
% Do not use $\backslash${\tt{section}} anymore after $\backslash${\tt{appendix}}, only $\backslash${\tt{section*}}.
% If you have multiple appendixes use $\backslash${\tt{appendices}} then use $\backslash${\tt{section}} to start each appendix.
% You must declare a $\backslash${\tt{section}} before using any $\backslash${\tt{subsection}} or using $\backslash${\tt{label}} ($\backslash${\tt{appendices}} by itself
%  starts a section numbered zero.)}


% \section{References Section}
% You can use a bibliography generated by BibTeX as a .bbl file.
%  BibTeX documentation can be easily obtained at:
%  http://mirror.ctan.org/biblio/bibtex/contrib/doc/
%  The IEEEtran BibTeX style support page is:
%  http://www.michaelshell.org/tex/ieeetran/bibtex/
 
%  % argument is your BibTeX string definitions and bibliography database(s)
% %\bibliography{IEEEabrv,../bib/paper}
% You can manually copy in the resultant .bbl file and set second argument of $\backslash${\tt{begin}} to the number of references
%  (used to reserve space for the reference number labels box).

\newpage

\bibliography{refs}
\bibliographystyle{IEEEtran}
% \nocite{*}

% \newpage

 

% \end{IEEEbiography}



\vfill

\end{document}